\documentclass[journal]{IEEEtran}
\usepackage{amsmath,amsthm}
\usepackage{amssymb}
\usepackage{amsfonts}
\usepackage{graphicx}
\graphicspath{{graphics/}}
\usepackage{tikz,tikz-3dplot}
\usepackage{tikzscale}
\usepackage{xcolor}
\usepackage{bm}
\usepackage{dsfont}
\usepackage{stfloats}
\usepackage[flushleft]{threeparttable}
\usepackage[numbers,sort&compress]{natbib}

\usepackage[switch]{lineno}

\DeclareMathOperator*{\argmin}{arg\,min}

\usepackage[colorlinks]{hyperref}
\hypersetup{colorlinks,breaklinks,linkcolor=blue,urlcolor=blue,anchorcolor=blue,citecolor=blue}
\usepackage{booktabs}
\usepackage{tabularx}
\usepackage{multirow}
\usepackage{lscape}
\usepackage{subcaption}
\usepackage{epstopdf}
\usepackage{booktabs}
\usepackage{caption}
\usepackage{enumitem}

\usepackage{float}
\usepackage{microtype}
\usepackage{placeins}
\usepackage{xcolor}

\usepackage{threeparttable}
\usepackage{dcolumn}
\newcolumntype{d}[1]{D{.}{.}{#1}}
\newcommand{\tensor}[1]{\boldsymbol{\mathcal{#1}}}
\newcommand{\mat}[1]{\boldsymbol{#1}}

\ifCLASSINFOpdf
\else
\fi

\usepackage{algorithmic}
\usepackage{algorithm}

\usepackage{xcolor}
\definecolor{darkblue}{rgb}{0.0,0.5,0.5}

\begin{document}
%



\title{Hankel-structured Tensor Robust PCA for Multivariate Traffic Time Series Anomaly Detection}


\author{\IEEEauthorblockN{Xudong Wang, Luis Miranda-Moreno, and Lijun Sun}


\thanks{This research is supported by the Natural Sciences  and Engineering Research Council (NSERC) of Canada, the Fonds de recherche du Quebec - Nature et technologies (FRQNT), and the Canada Foundation for Innovation (CFI). X. Wang would like to thank FRQNT for providing the B2X Doctoral Scholarship. }

\thanks{The authors are with the Department of Civil Engineering, McGill University, Montreal, Quebec H3A 0C3, Canada.
Corresponding author: L. Sun  (Email: lijun.sun@mcgill.ca)}
}



%



\IEEEtitleabstractindextext{%
\begin{abstract}

Spatiotemporal traffic data (e.g., link speed/flow) collected from sensor networks can be organized as multivariate time series with additional spatial attributes. A crucial task in analyzing such data is to identify and detect anomalous observations and events from the data with complex spatial and temporal dependencies. Robust Principal Component Analysis (RPCA) is a widely used tool for anomaly detection. However, the traditional RPCA purely relies on the global low-rank assumption while ignoring the local temporal correlations. In light of this, this study proposes a Hankel-structured tensor version of RPCA for anomaly detection in spatiotemporal data. We treat the raw data with anomalies as a multivariate time series matrix (location $\times$ time) and assume the denoised matrix has a low-rank structure. Then we transform the low-rank matrix to a third-order tensor by applying temporal Hankelization. In the end, we decompose the corrupted matrix into a low-rank Hankel tensor and a sparse matrix. With the Hankelization operation, the model can simultaneously capture the global and local spatiotemporal correlations and exhibit more robust performance. We formulate the problem as an optimization problem and use tensor nuclear norm (TNN) to approximate the tensor rank and $l_1$ norm to approximate the sparsity. We develop an efficient solution algorithm based on the Alternating Direction Method of Multipliers (ADMM). Despite having three hyper-parameters, the model is easy to set in practice. We evaluate the proposed method by synthetic data and metro passenger flow time series and the results demonstrate the accuracy of anomaly detection.

\end{abstract}

}

\maketitle

\IEEEdisplaynontitleabstractindextext

%
\IEEEpeerreviewmaketitle




\section{Introduction}
With the advances in information and communication technologies (ICT), large-scale spatiotemporal sensing data are being collected in various domains, such as climate science, environmental monitoring, traffic operation \cite{sofuoglu2020gloss}. The massive spatiotemporal data provide us with an excellent opportunity to understand the underlying patterns and dynamics of the system by exploiting the inherent spatiotemporal correlations and dependency structures. Anomaly detection is one of the crucial tasks in spatiotemporal analysis, which aims to distinguish unusual phenomena/behaviors/events from regular ones. For example, in intelligent transportation systems (ITS), anomaly has different definitions in terms of applications: it can be the spike in passenger flow, the congestion of the road, or the change of travel behavior. Detecting anomalies can help transportation agencies understand how the system performs and provide helpful information to timely adjust the traffic control/management plans \cite{wang2021diagnosing}.


Many existing studies apply unsupervised learning methods to detect anomalies since they can address the imbalanced and unlabeled data in a more principled manner \cite{zhang2021unsupervised}. Robust Principal Component Analysis (RPCA)---a simple and non-parametric method---is a widely applied technique on this track \cite{wang2018improved,jin2017sparse,wei2021metro}. Given a corrupted matrix $\mat{M} \in \mathbb{R}^{N\times T}$ ($N$ denotes locations/sensors and $T$ denotes time steps in spatiotemporal traffic flow matrix), the purpose of RPCA is to decompose the $\mat{M}$ into a low-rank matrix (true data) $\mat{L}\in \mathbb{R}^{N\times T}$ and a sparse matrix (anomalies) $\mat{S}\in \mathbb{R}^{N\times T}$. The RPCA can be formulated as a nonconvex optimization problem \cite{candes2011robust},
\begin{equation}
    \label{eq:rpca}
    \underset{\mat{L,S}}{\min}~ \text{rank}(\mat{L})+\gamma \|\mat{S}\|_0, \quad \text{s.t.} \quad \mat{L} + \mat{S} =  \mat{M},
\end{equation}
where $\text{rank}(\cdot)$ measures the matrix rank, $\|\cdot\|_0$ denotes $l_0$ norm used to count the number of nonzero entries in the matrix, and $\gamma>0$ is a trade-off parameter. In the literature, problem \eqref{eq:rpca} can be solved by either convex relaxations---approximating the matrix rank by nuclear norm and the $l_0$ norm by $l_1$ norm under certain conditions \cite{candes2011robust, cai2010singular}, or nonconvex methods, such as matrix factorization \cite{wang2012probabilistic,jin2017sparse} and alternating minimization \cite{zhang2019correction,cai2021accelerated}. More references about RPCA solution can be found in \cite{ma2018efficient}. It should be noted that the RPCA problem is converted to robust matrix completion (RMC) problem if $\mat{M}$ is not fully observed, and matrix completion (MC) problem is a special case of RMC when $\mat{S}=\boldsymbol{0}$ \cite{zhang2019correction}.






Despite that RPCA has achieved success in various applications, such as image-inpainting, face recognition and background subtraction \cite{candes2011robust}, the model has a fundamental limitation when deal with  spatiotemporal data. RPCA only relies on the matrix structure and ignores the spatiotemporal correlation/dependency of the data. For instance, the result of the optimization model is invariant to the perturbation of columns (i.e., time) in $\mat{M}$. As a result, the model has a risk of overfitting to noise and degrades the performance of detecting valid anomalies. {One way to address this issue is to integrate additional regularizers to impose temporal consistency.} For example, \citet{wang2018improved} and  \citet{wei2021metro} included Toeplitz temporal regularizer to ensure the observations from adjacent timestamps to be similar. In \citet{chen2021low}, temporal variation is introduced as a generative approach to ensure each time series follows parametric autoregressive (AR) models. These regularization-based methods have shown superior performance in modeling corrupted spatiotemporal data. However, the use of Toeplitz or AR regularizers is limited to their function forms; they are often insufficient to characterize the complex temporal patterns. Moreover, the additional regularizer also brings new weight parameters to the model, which makes the parameter tuning process more difficult. {Another way is to augment the spatiotemporal information in data structure directly without adding extra temporal regularizer to the model. For example,  \citet{li2013efficient} proposed a probabilistic principal
component analysis (PPCA) based imputing method by incorporating one-step delay between adjacent points.}

\begin{figure*}[!t]
    \centering
    \includegraphics{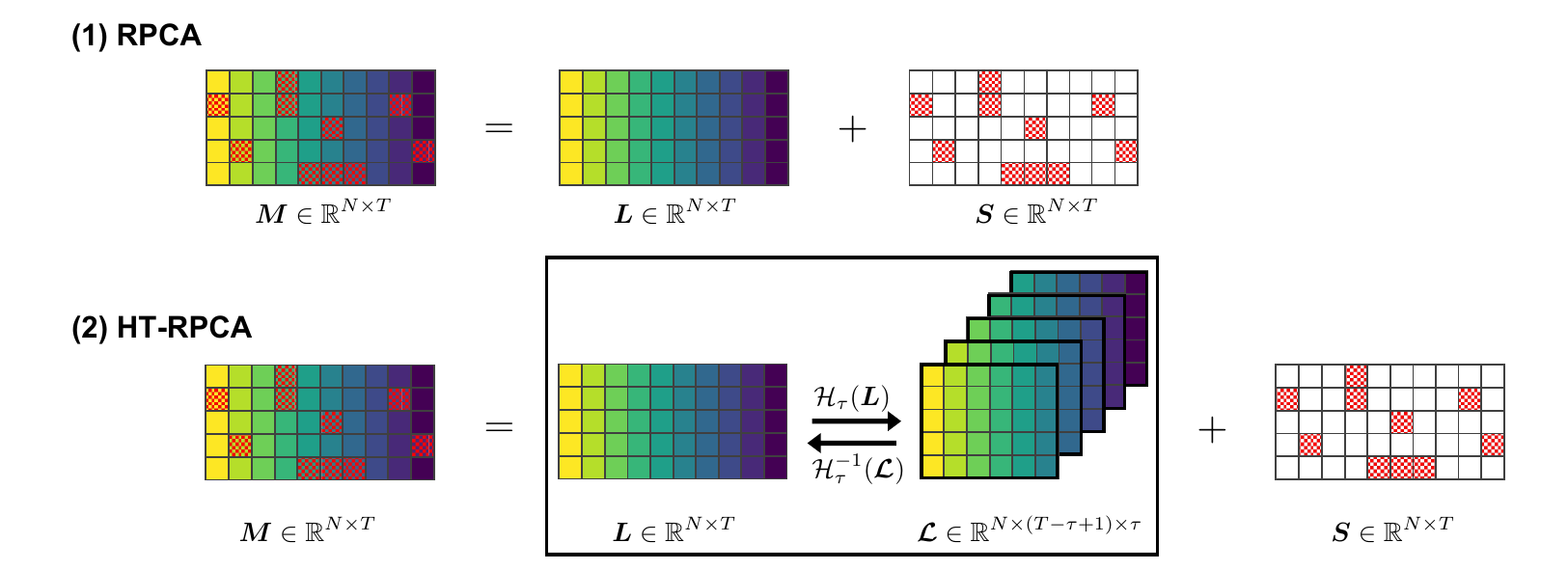}
    \caption{Illustrations of RPCA and the proposed Hankel-structured Tensor RPCA (HT-RPCA) for spatiotemporal data collected from $N$ stations during $T$ timestamps. Cell with red patch represent anomalous data. The Hankelization operator $\mathcal{H}_\tau$ and the inverse Hankelization operator $\mathcal{H}_\tau^{-1}$ is introduced in Section \ref{sec:hankel}. }
    \label{fig:rpca}
\end{figure*}

In this study, we develop a new RPCA-based model for anomaly detection for multivariate traffic time series. Motivated by the fact that traffic data often exhibit strong periodic and quasiperiodic patterns (e.g., day-to-day similarity) \cite{li2015trend}, we suggest to use Hankelization---a natural data augmentation technique for time series data---on the traffic data to incorporate the intrinsic temporal correlation. The delay embedding length $\tau$ in the Hankelization process can be set to the maximum periodicity in the signal. In light of this, the $N\times T$ matrix $\mat{L}$ can be transferred to a third-order tensor $\tensor{L} \in \mathbb{R}^{N\times(T-\tau+1)\times \tau}$. By doing so, the Hankel tensor can preserve the global pattern of the low-rank data and introduce a higher-order dependency/correlation structure within a local temporal domain \cite{wang2021lowrank}. Figure~\ref{fig:rpca} illustrates the vanilla RPCA model and the proposed Hankel-structured tensor RPCA  (HT-RPCA) model, respectively. We solve the HT-RPCA model by the Alternating Direction Method of Multipliers (ADMM). Specifically, we use tensor nuclear norm (TNN) to approximate the tensor rank and $l_1$ norm to approximate the sparsity. To be more general, we also extend it to Hankel-structured tensor RMC model (HT-RMC) to detect anomalies from partially observed data (i.e., with missing values). We first design a synthetic dataset to quantify the performance of the proposed models by the root mean square error (RMSE) and the absolute error (MAE), then we test the model on metro passenger flow data collected from Guangzhou, China, to detect the unusual passenger flow at station-level and analyze the anomaly propagation through the metro network.




The idea of leveraging Hankel matrix/tensor has been introduced in some recent studies in different areas, and it is a key component in the singular spectrum analysis (SSA) of time series data and signals \citep{golyandina2001analysis}. Recently, Hankelization has also been integrated with RPCA/RMC/MC to model multivariate time series data and multi-channel signals. For example, \citet{zhang2019correction} developed an alternating-projection-based algorithm to solve the RMC by introducing the Hankel matrix structure. In the work of \cite{jin2017sparse}, the authors incorporated the Hankel matrix and used matrix factorization to relax the nuclear norm of the Hankel matrix. For the Hankel-structured tensor, \citet{kasai2016network} proposed an online Hankel tensor model based on the CANDECOMP/PARAFAC (CP) decomposition to infer network-level anomalies from indirect link measurements. The closest framework to our work is \citet{xu2021fast} with an application on visual inpainting. However, the model targets on tensor completion problem instead of anomaly detection. In a nutshell, the main contributions of our work are summarized as follows:
\begin{itemize}[noitemsep]
    \item We model the spatiotemporal data from $N$ locations or sensors over $T$ time steps as a  multivariate time series matrix and apply temporal delay embedding to transfer the matrix into a third-order Hankel tensor. In doing so, both the global and local spatiotemporal correlations can be incorporated naturally in the tensor.
    \item We propose a Hankel-structured tensor RPCA model to handle the anomaly detection problem from spatiotemporal matrix. Specifically, we use tensor nuclear norm (TNN) to approximate tensor rank and $l_1$ norm to approximate number of anomalous observations. The model can be solved efficiently by using the ADMM framework.
    \item We evaluate the performance of HT-RPCA on two dataset: a synthetic data set and a metro passenger flow data set. The results show that the proposed method exhibits superior performance on anomaly detection compared with the baseline models.
\end{itemize}

The remainder of this paper is organized as follows. Section ~\ref{sec:notations} introduces the notations and preliminaries about tensor nuclear norm. In Section~\ref{sec:method}, we introduce the HT-RPCA and HT-RMC model in detail and develop an ADMM algorithm for model estimation. In Section~\ref{sec:casestudy}, we apply the synthetic data and metro passenger flow to evaluate the performance of the proposed method. Section~\ref{sec:conclusion} concludes this study and discusses some directions for future research.


\section{Notations and Preliminaries}
\label{sec:notations}
\subsection{Notations}

In this paper, we use lowercase letters to denote scalars, e.g., $x \in \mathbb{R}$, boldface lowercase letters to denote vectors, e.g., $\boldsymbol{x}\in \mathbb{R}^{n}$, boldface capital letters to denote matrices, e.g., $\boldsymbol{X}\in \mathbb{R}^{n_1 \times n_2}$, and boldface Euler script letters to denote third-order tensors, e.g., $\tensor{X} \in \mathbb{R}^{n_1 \times n_2 \times n_3}$. We denote the $(i,j)$th entry of a matrix by $\mat{X}_{i,j}$ or $x_{ij}$ and the $(i,j,k)$th entry of a third-order tensor by $\tensor{X}_{i,j,k}$ or $x_{ijk}$. We use the MATLAB notation $\tensor{X}(i,:,:)$, $\tensor{X}(:,i,:)$ and $\tensor{X}(:,:,i)$ to denote the $i$th horizontal, lateral and frontal slices of a tensor, respectively. For simplicity, the frontal slice $\tensor{X}(:,:,i)$ is denoted by $\mat{X}^{(i)} \in \mathbb{R}^{n_1\times n_2}$.

Given a matrix $\mat{X} \in \mathbb{R}^{n_1 \times n_2}$, the $l_1$-norm is denoted as $\|\mat{X}\|_1=\sum_{i=1}^{n_1}\sum_{j=1}^{n_2}|{x}_{ij}|$, the nuclear norm (NN) is denoted as $\|\boldsymbol{X}\|_*=\sum_{i=1}^{\min(n_1,n_2)} \sigma_i(\boldsymbol{X})$, where $\sigma_i(\boldsymbol{X})$ is the $i$th largest singular value of $\boldsymbol{X}$ \cite{liu2012tensor}, and the Frobenius norm is defined as $\|\mat{X}\|_F = \sqrt{\sum_{i=1}^{n_1}\sum_{j=1}^{n_2} {x}_{ij}^2}$. The inner product of two matrices of the same size is $\left<\mat{X}, \mat{Y}\right>=\sum_{i=1}^{n_1} \sum_{j=1}^{n_2} {x}_{ij}{y}_{ij}$.

\subsection{Preliminaries}

For a third-order tensor $\tensor{A} \in \mathbb{R}^{n_1\times n_2 \times n_3}$, we define the unfold operation that maps $\tensor{A}$ to a matrix of size $n_1n_3\times n_2$ and its inverse fold operator as:
\begin{equation*}
    \text{unfold}(\tensor{A}) = \left[ \begin{array}{c}
         \mat{A}^{(1)}  \\
         \mat{A}^{(2)} \\
         \vdots \\
         \mat{A}^{(n_3)}
    \end{array}\right],
    \quad \text{fold}(\text{unfold}(\tensor{A}))=\tensor{A}.
\end{equation*}

We denote the block circulant matrix from frontal slices of $\tensor{A}$ as $\text{bcirc}(\tensor{A}) \in \mathbb{R}^{n_1 n_3 \times n_2n_3}$
\begin{equation*}
    \text{bcirc}(\tensor{A})=\left[ \begin{array}{cccc}
         \mat{A}^{(1)}& \mat{A}^{(n_3)}& \cdots & \mat{A}^{(2)}  \\
         \mat{A}^{(2)}& \mat{A}^{(1)}& \cdots & \mat{A}^{(3)}  \\
         \vdots& \vdots& \ddots&\vdots \\
         \mat{A}^{(n_3)}& \mat{A}^{(n_3-1)}& \cdots & \mat{A}^{(1)}  \\
    \end{array}\right].
\end{equation*}

\vspace{0.5em}
\noindent\textbf{Definition 1 Tensor product (t-product) \cite{kilmer2011factorization}:} \textit{Let $\tensor{A} \in \mathbb{R}^{n_1\times n_2 \times n_3}$ and $\tensor{B} \in \mathbb{R}^{n_2\times l \times n_3}$, then the t-product $\tensor{A} \ast \tensor{B}$ is defined to be a tensor of size $n_1 \times l \times n_3$,}
\begin{equation}
\label{eq:tpro}
    \tensor{A} \ast \tensor{B} = \text{fold}(\text{bcirc}(\tensor{A})\cdot \text{unfold}(\tensor{B})).
\end{equation}

The t-product defined in \eqref{eq:tpro} is usually calculated in the Fourier domain by discrete Fourier transformation (DFT) as it is cumbersome to apply circular convolution between the elements in the time domain. Let $\tensor{\bar{A}} \in \mathbb{C}^{n_1 \times n_2 \times n_3}$ denote as the DFT result on $\tensor{A}$ along the third dimension. Following the MATLAB command fft and ifft, we have
\begin{equation*}
    \tensor{\bar{A}} = \text{fft}(\tensor{A},[~],3), \quad \tensor{{A}} = \text{ifft}(\tensor{\bar{A}},[~],3).
\end{equation*}

We denote $\mat{\bar{A}}^{(i)} \in \mathbb{C}^{n_1 \times n_2}$ as the $i$th frontal slice of $\tensor{\bar{A}}$ and $\mat{\bar{A}} \in \mathbb{C}^{n_1 n_3 \times n_2 n_3}$ as a block diagonal matrix with $\mat{\bar{A}}^{(i)}$:
\begin{equation*}
    \mat{\bar{A}}=\left[ \begin{array}{cccc}
         \mat{\bar{A}}^{(1)}& &  &   \\
         & \mat{\bar{A}}^{(2)}& &   \\
         & & \ddots& \\
         & &  & \mat{\bar{A}}^{(n_3)}  \\
    \end{array}\right].
\end{equation*}
In light of this, the t-product is equivalent to the matrix multiplication in the Fourier domain, i.e., $ \tensor{C} = \tensor{A} \ast \tensor{B} \Leftrightarrow \mat{\bar{C}} = \mat{\bar{A}}\mat{\bar{B}}$ \cite{lu2019tensor}.

\vspace{0.5em}
\noindent\textbf{Definition 2 Tensor SVD (t-SVD) \cite{kilmer2011factorization}:} \textit{Let $\tensor{A} \in \mathbb{R}^{n_1 \times n_2 \times n_3}$, then it can be factorized as}
\begin{equation}
    \tensor{A} = \tensor{U} \ast \tensor{S} \ast \tensor{V}^*,
\end{equation}
\textit{where $\tensor{U} \in \mathbb{R}^{n_1 \times n_1 \times n_3}$, $\tensor{V} \in \mathbb{R}^{n_2 \times n_2 \times n_3}$ are orthogonal, and $\tensor{S} \in \mathbb{R}^{n_1 \times n_2 \times n_3}$ is an f-diagonal tensor. $\tensor{V}^*$ is the conjugate transpose of $\tensor{V}$.}

\vspace{0.5em}
\noindent\textbf{Definition 3 Tensor nuclear norm (TNN) \cite{lu2019tensor}:} \textit{Let $ \tensor{A} = \tensor{U} \ast \tensor{S} \ast \tensor{V}^*$ be the t-SVD of $\tensor{A} \in \mathbb{R}^{n_1 \times n_2 \times n_3}$. The tensor nuclear norm of $\tensor{A}$ is defined as }
\begin{equation}
\label{eq:tnn}
    \|\tensor{A}\|_* := \left<\tensor{S}, \tensor{I}\right> = \sum_{i=1}^r\tensor{S}(i,i,1),
\end{equation}
\textit{where $r=\text{rank}_t(\tensor{A})$ is the tubal rank}.


\section{Methodology}
\label{sec:method}

In this section, we first introduce the temporal Hankelization process to transform a spatiotemporal matrix to a third-order Hankel tensor and the corresponding inverse operation. Then we propose a Hankel-structured tensor RPCA (HT-RPCA) model and RMC model (HT-RMC) to detect anomalies and present the solution algorithm based on ADMM.

\subsection{Temporal Hankel Tensor Transformation}
\label{sec:hankel}

The temporal Hankelization operator $\mathcal{H}_{{\tau}}$ with temporal delay embedding length $\tau$ transforms a given spatiotemporal matrix $\mat{X}\in \mathbb{R}^{N\times T}$ to a third-order Hankel tensor $\tensor{X}=\mathcal{H}_{{\tau}}\left(\mat{X}\right)\in \mathbb{R}^{ N \times (T-\tau+1) \times \tau}$. In the Hankel tensor we have
\begin{equation}
    \tensor{X}_{:,:,t} = \mat{X}_{:,t:t+T-\tau+1} \in \mathbb{R}^{ N \times (T-\tau+1)}, ~ t=1,\ldots, \tau.
\end{equation}

\begin{figure*}[!t]
    \centering
    \includegraphics{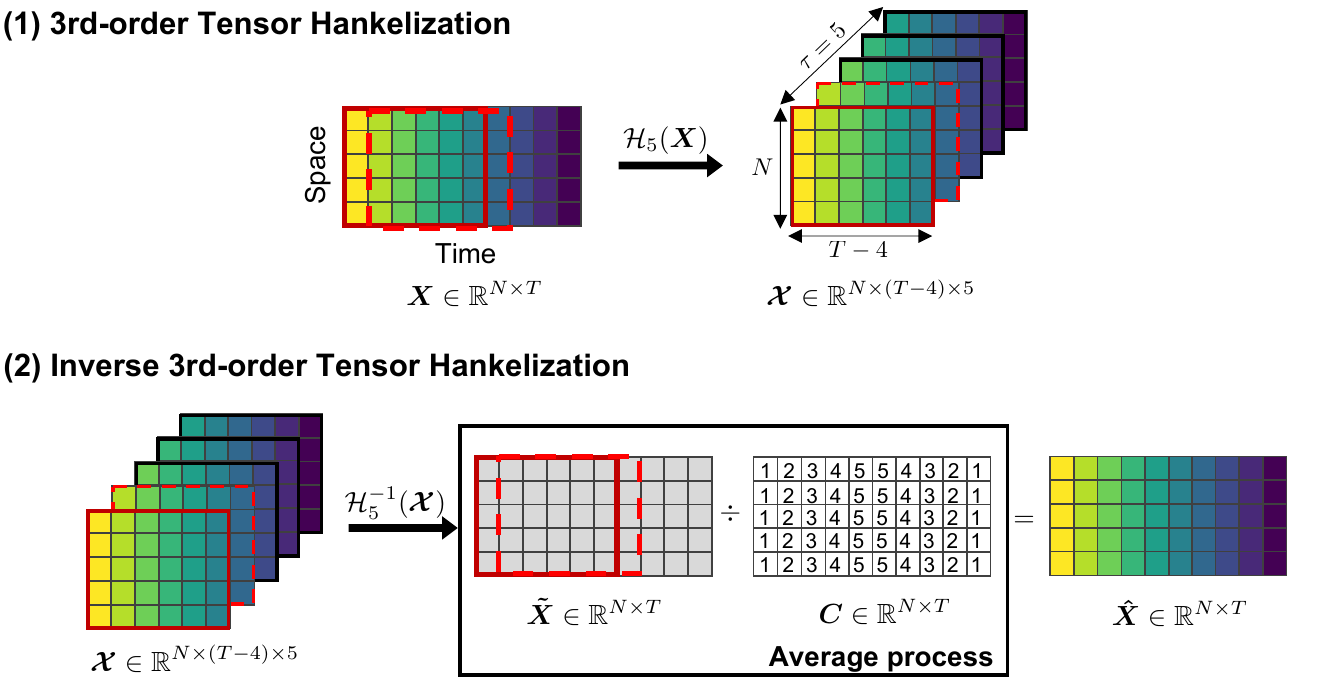}
    \caption{An example of Hankelization $\mathcal{H}_\tau$ and inverse Hankelization $\mathcal{H}_\tau^{-1}$ with $\tau=5$ for a matrix $\mat{X} \in \mathbb{R}^{N \times T}$.}
    \label{fig:hankel}
\end{figure*}

Correspondingly, the inverse Hankelization operation $\mathcal{H}_{{\tau}}^{-1}$ is to transform a Hankel tensor $\tensor{X}$ to a matrix $\mat{\hat{X}}$ by averaging the corresponding entries in the Hankel tensor \cite{wang2021lowrank}. In other words, we aggregate the sub-matrix $\tensor{X}_{:,:,t}$ for $t=1,\cdots,\tau$ along temporal dimension by shifting one step each time to obtain $\mat{\Tilde{X}}$ and then we divide the corresponding repeat times of each entry. Figure~\ref{fig:hankel} illustrates an example of the tensor Hankelization and the inverse tensor Hankelization with $\tau=5$. The count matrix $\mat{C}$ represents the repetition number of each entry appeared in the Hankel tensor.

\subsection{Hankel-structured Tensor of RPCA (HT-RPCA)}

We denote the spatiotemporal data collected from $N$ locations/sensors over $T$ timestamps by $\mat{Z} \in \mathbb{R}^{N \times T}$ and introduce an auxiliary matrix $\mat{M}$ ($\mat{M} = \mat{Z}$). We assume the complete and corrupted matrix $\mat{M}\in \mathbb{R}^{N \times T}$ can be decomposed into a low-rank matrix with Hankel constraint $\mat{L} = \mathcal{H}_\tau^{-1}(\tensor{L})$, where $\tensor{L}\in\mathbb{R}^{N \times (T-\tau+1) \times \tau}$ is the low-rank Hankel tensor, and a sparse matrix $\mat{S}$. With the temporal Hankelization process, we transform the matrix-based problem to a tensor-based problem. The HT-RPCA model can be described as the following optimization problem:
\begin{equation}
\begin{aligned}
\label{eq:hankeltnn}
     &\min_{\tensor{L}, \mat{S}}~  \text{rank}(\tensor{L}) + \gamma \|\mat{S}\|_0,\\
& \text{s.t.}~\left\{\begin{array}{l} \tensor{L}=\mathcal{H}_{{\tau}}\left(\mat{L}\right), \\
\mat{L+S=M}.\\
  \end{array} \right.
\end{aligned}
\end{equation}

Same as problem \eqref{eq:rpca}, computing the tensor rank and $l_0$ norm in \eqref{eq:hankeltnn} is an NP-hard problem. To address the problem, the tensor rank can be approximated by convex relaxation, such as the sum of nuclear norms (SNN) of all tensor unfoldings \cite{liu2012tensor}, the squared nuclear norm (SqNN) of a balanced tensor unfolding \cite{mu2014square}, the tensor nuclear norm (TNN) \cite{lu2019tensor} and its variants \cite{gao2020enhanced,wang2020robust}, or by nonconvex methods like tensor factorization \cite{kasai2016network}.

In our model, we use TNN to approximate the tensor rank as it is a tight convex relaxation of the tensor average rank \cite{lu2019tensor} and $l_1$ norm to approximate matrix sparsity. In light of this, the noncovex problem \eqref{eq:hankeltnn} can be solved by:
\begin{equation}
\begin{aligned}
\label{eq:HT-RPCA}
     &\min_{\tensor{L}, \mat{S}}~ \| \tensor{L}\|_* + \gamma \|\mat{S}\|_1,\\
& \text{s.t.}~\left\{\begin{array}{l} \tensor{L}=\mathcal{H}_{{\tau}}\left(\mat{L}\right), \\
\mat{L+S=M}.\\
  \end{array} \right.
\end{aligned}
\end{equation}
where $\|\cdot\|_*$ is the TNN defined in Eq.~\eqref{eq:tnn}.

The convex optimization problem of \eqref{eq:HT-RPCA} can be efficiently solved by using the ADMM framework. The augmented Lagrangian function is
\begin{equation}
\begin{aligned}
\label{eq:lag}
\mathcal{L}\left(\tensor{L},\mat{S},\mat{E}\right) & =  \|\tensor{{L}}\|_* + \gamma \|\mat{S}\|_1 \\
&+ \frac{\rho}{2}\|\mat{M} - \mathcal{H}_{\tau}^{-1}({\tensor{L}})-\mat{S}\|_{F}^{2}\\
&+ \left\langle \mat{M}-\mathcal{H}_{\tau}^{-1}(\tensor{L})-\mat{S},\mat{E}\right\rangle, \\
\end{aligned}
\end{equation}
where $\rho >0$ is a penalty parameter and $\mat{E}$ is a dual variable. With the augmented Lagrangian function Eq.~\eqref{eq:lag}, the optimization problem \eqref{eq:HT-RPCA} can be solved iteratively. The inferences of variables $\tensor{L}$, $\mat{S}$ and $\mat{E}$ at the $\ell$th iteration are given below.

\vspace{0.5em}
\textit{1) Update variable $\tensor{L}$:}
\begin{equation}
    \label{eq:admm_l}
    \begin{aligned}
        \tensor{L}^{\ell+1} : =  ~&\underset{\tensor{L}}{\arg\min~} \frac{1}{\rho} \| \tensor{{L}}^\ell\|_*\\
        + &\frac{1}{2}\| \tensor{L}^\ell- \mathcal{H}_{\tau}\left(\mat{M}^\ell - \mat{S}^\ell + \frac{1}{\rho^\ell}\mat{E}^\ell\right)\|_F^2.\\
        =~&\mathcal{D}_{1/\rho}\left(\mathcal{H}_\tau\left(\mat{M}^\ell - \mat{S}^\ell + \frac{1}{\rho^\ell}\mat{E}^\ell\right)\right),
    \end{aligned}
\end{equation}
where $\mathcal{D}_{\cdot}(\cdot)$ denotes the tensor singular value thresholding (t-SVT) as shown in Lemma 1.

\vspace{0.5em}
\noindent\textbf{Lemma 1}: \textit{Let $\tensor{Y} = \tensor{U}*\tensor{S}*\tensor{V}$ be the tensor SVD of $\tensor{Y} \in \mathbb{R}^{n_1 \times n_2 \times n_3}$. For any $\tau>0$, the tensor singular value thresholding operator $\mathcal{D}_\tau$ obeys}
\begin{equation}
    \mathcal{D}_\lambda(\tensor{Y}) = \underset{\tensor{X}\in \mathbb{R}^{n_1 \times n_2 \times n_3}}{\arg\min~} \lambda \|\tensor{X}\|_* + \frac{1}{2}\|\tensor{X}-\tensor{Y}\|_F^2,
\end{equation}
\textit{where $\mathcal{D}_{\lambda}(\tensor{Y}) = \tensor{U}*\tensor{S}_\lambda*\tensor{V}$} and $\tensor{S}_\lambda = \text{ifft}((\max(\tensor{\bar{S}}-\lambda),0),[~],3)$ \cite{lu2019tensor}. Algorithm \ref{alg:SVT} summarizes the details of computing $\mathcal{D}_{\lambda}(\tensor{Y})$ for a third-order tensor.
\begin{algorithm}
\textbf{Input}: $\tensor{Y} \in \mathbb{R}^{n_1 \times n_2 \times n_3}$\\
\textbf{Output}: $\mathcal{D}_{\lambda}(\tensor{Y})$
    \begin{algorithmic}[1]
    \STATE $\tensor{\bar{Y}} = \text{fft}(\tensor{Y},[~~],3)$.
    \FOR{$i=1,\dots, \lceil\frac{\tau+1}{2}\rceil$}
    \STATE $[\mat{U,S,V}] = \text{SVD}(\mat{\bar{Y}}^{(i)})$;
    \STATE $\mat{\bar{W}}^{(i)} = \mat{U}\cdot \max((\mat{S}-\lambda),0) \cdot \mat{V}^*$.
    \ENDFOR
    \FOR{$i= \lceil\frac{\tau+1}{2}+1\rceil, \dots, \tau$}
    \STATE $\mat{\bar{W}}^{(i)} = \text{conj}(\mat{\bar{W}}^{(\tau-i+2)})$.
    \ENDFOR
    \STATE $\mathcal{D}_\lambda(\tensor{Y})=\text{ifft}(\tensor{\bar{W}},[~~],3)$.
    \end{algorithmic}
    \caption{Tensor Singular Value Thresholding (t-SVT) \cite{lu2019tensor}}
    \label{alg:SVT}
\end{algorithm}

\vspace{0.5em}
\textit{2) Update variable $\mat{S}$:}
\begin{equation}
    \label{eq:admm_s}
    \begin{aligned}
        \mat{S}^{\ell+1} :&= ~ \underset{\mat{S}}{\argmin~}  \frac{\gamma}{\rho}\|\mat{S}^\ell\|_1\\
        +&\frac{1}{2}\| \mat{S}^\ell-\left(\mat{M}^\ell -  \mathcal{H}_{\tau}^{-1}(\tensor{L}^{\ell+1})+\frac{1}{\rho^\ell}\mat{E}^{\ell}\right)\|_F^2,\\
        =& ~\mathcal{S}_{\gamma/\rho}\left(\mat{M}^\ell -  \mathcal{H}_{\tau}^{-1}(\tensor{L}^{\ell+1})+\frac{1}{\rho^\ell}\mat{E}^{\ell}\right).
    \end{aligned}
\end{equation}
where $\mathcal{S}_{\cdot}(\cdot)$ denotes the soft shrinkage operator introduced in Lemma 2.

\vspace{0.5em}
\noindent\textbf{Lemma 2}: \textit{For any $\tau>0$ and $\mat{Y} \in \mathbb{R}^{n_1 \times n_2}$, the soft shrinkage operator $\mathcal{S}_\tau$ obeys}
\begin{equation}
    \mathcal{S}_\lambda(\mat{Y}) = \underset{\mat{X}\in \mathbb{R}^{n_1 \times n_2}}{\arg\min~} \lambda \|\mat{X}\|_1 + \frac{1}{2}\|\mat{X}-\mat{Y}\|_F^2,
\end{equation}
\textit{where} $\mathcal{S}_\lambda(\mat{Y})=\text{sgn}(\mat{Y})\max\left(|\mat{Y}|-\lambda,0\right)$ \cite{candes2011robust}.

\vspace{0.5em}
\textit{3) Update variable $\mat{E}$:}

The dual variable $\mat{E}$ is updated by
\begin{equation}
\label{eq:e_sol}
    \mat{E}^{\ell+1} = \mat{E}^{\ell} + \rho^\ell \left( \mat{M} - \mathcal{H}_{\tau}^{-1}(\tensor{L}^{\ell+1}) - \mat{S}^{\ell+1}\right),
\end{equation}
where $\rho^{\ell+1} = \beta \rho^\ell$ with $\beta \in [1.0, 1.2]$ to accelerate the algorithm \cite{oh2017fast}.

\subsection{Hankel-structured Tensor of RMC (HT-RMC)}

In practice, the observations $\mat{Z}$ might be incomplete due to various reasons, such as sensor disfunction and signal communication failure. We denote the incomplete matrix as $\mathcal{P}_\Omega(\mat{Z})$, where the operator $\mathcal{P}_\Omega: \mathbb{R}^{N \times T} \rightarrow \mathbb{R}^{N \times T}$ is defined as $[\mathcal{P}_\Omega(\mat{Z})]_{i,j}=\mat{Z}_{i,j}$, if $(i,j)\in\Omega$, and $[\mathcal{P}_{\bar{\Omega}} (\mat{Z})]_{i,j}=0$ otherwise. In light of this, we propose an RMC model with Hankel-structured tensor (HT-RMC) to detect anomalies from the partial observations and complete the missing values simultaneously. Based on \eqref{eq:HT-RPCA}, the objective function of HT-RMC is easy to obtain with incorporating partial observations constraint:
\begin{equation}
\begin{aligned}
\label{eq:HT-RMC}
     &\min_{\tensor{L}, \mat{S}}~ \| \tensor{L}\|_* + \gamma \|\mat{S}\|_1,\\
& \text{s.t.}~\left\{\begin{array}{l} \tensor{L}=\mathcal{H}_{{\tau}}\left(\mat{L}\right), \\
\mat{L+S=M},\\
\mathcal{P}_\Omega(\mat{M}) = \mathcal{P}_\Omega(\mat{Z}).\\
  \end{array} \right.
\end{aligned}
\end{equation}

In the HT-RMC model, the update of $\tensor{L}$, $\mat{S}$ and $\mat{E}$ are the same as in HT-RPCA model. The variable $\mat{M}$ can be updated by solving
\begin{equation}
    \label{eq:admm_m}
    \begin{aligned}
        \mat{M}^{\ell+1} := ~ &\underset{\mat{M}}{\argmin~}  \frac{\rho}{2}\|\mat{M}^\ell - \mathcal{H}_{\tau}^{-1}({\tensor{L}^{\ell+1}})-\mat{S}^{\ell+1}\|_{F}^{2}\\
&+ \left\langle \mat{M}^{\ell}-\mathcal{H}_{\tau}^{-1}(\tensor{L}^{\ell+1})-\mat{S}^{\ell+1},\mat{E}^{\ell}\right\rangle.
    \end{aligned}
\end{equation}

Let the partial gradient of the optimal function Eq.~\eqref{eq:admm_m} equal to 0, the update equation of $\mat{M}^{\ell+1}$ is:
\begin{equation}
\label{eq:m_sol}
    \begin{aligned}
    &\mathcal{P}_{\bar{\Omega}}(\mat{M}^{\ell+1}) = \left(\mathcal{H}_{\tau}^{-1}(\tensor{L}^{\ell+1}) + \mat{S}^{\ell+1} - \frac{1}{\rho^\ell}\mat{E}^{\ell} \right)_{\bar{\Omega}},\\
    &\mathcal{P}_\Omega(\mat{M}^{\ell+1}) = \mathcal{P}_\Omega(\mat{Z}).
    \end{aligned}
\end{equation}
It can be seen that only missing values are estimated at each iteration, while the observed values are fixed. Therefore, the $\mat{M}$ always equals to the observation matrix $\mat{Z}$ in HT-RPCA model.

To stop the algorithm, the following convergence criterion is applied in the model:
 \begin{equation}
    \frac{||\mathcal{P}_\Omega(\mat{M}) - \mathcal{H}_{\tau}^{-1}(\tensor{L}^{\ell+1}) - \mat{S}^{\ell+1}||_F}{||\mathcal{P}_\Omega(\mat{M})||_F} < {tol},
\end{equation}
where \textit{tol} is the stopping threshold. Algorithm \ref{alg:HT-RPCA} summarizes the overall solution of HT-RMC. The algorithm is also HT-RPCA if we ignore the update of $\mat{M}$ (line 5 in Algorithm \ref{alg:HT-RPCA}).

\begin{algorithm}[htpb]
\textbf{Input}: $\mat{Z}_\Omega \in \mathbb{R}^{N \times T}$, $\tau$, \textit{tol}, $\rho$, $\rho_{\max}$ and $\beta$\\
\textbf{Output}: ${\mat{L}}$ and $\mat{S}$\\
\textbf{Initialize}: $\ell = 1$
    \begin{algorithmic}[1]
    \STATE $\mathcal{P}_\Omega(\mat{M})^\ell = \mathcal{P}_\Omega(\mat{Z})$ and $\mathcal{P}_{\bar{\Omega}}(\mat{M})^\ell = 0$
    \WHILE{not converged}
    \STATE Update $\tensor{L}$ by Eq.~\eqref{eq:admm_l}
    \STATE Update $\mat{S}$ by Eq.~\eqref{eq:admm_s}
    \STATE Update $\mat{M}$ by Eq.~\eqref{eq:m_sol}
    \STATE Update $\mat{E}$ by Eq.~\eqref{eq:e_sol}
    \STATE $\rho^{\ell+1} = \min(\beta\rho^\ell, \rho_{\max})$
    \STATE $\ell = \ell + 1$
    \ENDWHILE
    \end{algorithmic}
    \caption{Hankel-structured tensor RMC (HT-RMC)}
    \label{alg:HT-RPCA}
\end{algorithm}

Despite of having three hyper-parameters---the Hankel delay embedding length $\tau$, the low rank and sparse trade-off parameter $\gamma$, and the penalty parameter $\rho$, the proposed model is easy to tune in practice. The only important one is the balance parameter $\gamma$ which has to be tuned for each specific application, while  $\tau$ can be set to the maximum periodicity of the data and $\rho$ is usually set to a small number by default (e.g., $1\times 10^{-5}$) as it can be updated at each iteration.

\section{Case Study}  \label{sec:casestudy}

In this section, we compare the proposed model with four baseline models on both synthetic time-series data and real-world traffic data to evaluate the anomaly detection performance.

\subsection{Baseline Models}
\label{sec:baseline}
We compare the proposed model with the following baselines in the experiments. The objective function is also given in each baseline.

\begin{itemize}
    \item Robust principal component analysis (RPCA) \cite{candes2011robust}:
    \begin{equation*}
        \underset{\mat{L,S}}{\min}~ \|\mat{L}\|_*+\gamma \|\mat{S}\|_1, \quad \text{s.t.} \quad \mat{L} + \mat{S} =  \mat{M}.
    \end{equation*}

    \item RPCA with Toeplitz temporal regularizer $\mat{R}_{TV}$ (RPCA-TV) \cite{wang2018improved}:
    \begin{equation*}
    \begin{aligned}
        &\underset{\mat{L,S,P,U,V}}{\min}~ \|\mat{UV}\|_*+\lambda_1 \|\mat{S}\|_1+\lambda_2\|\mat{P}\|_1, \\ &\text{s.t.} \quad \mat{L} + \mat{S} =  \mat{M},~ \mat{L} = \mat{UV},~ \mat{R}_{TV}\mat{L}=\mat{P}.
    \end{aligned}
    \end{equation*}

    \item RPCA with periodic temporal regularizer $\mat{R}_{\tau}$ (RPCA-$\tau$).
    \begin{equation*}
    \begin{aligned}
        &\underset{\mat{L,S,P,U,V}}{\min}~ \|\mat{UV}\|_*+\lambda_1 \|\mat{S}\|_1+\lambda_2\|\mat{P}\|_1, \\ &\text{s.t.} \quad \mat{L} + \mat{S} =  \mat{M},~ \mat{L} = \mat{UV},~ \mat{R}_{\tau}\mat{L}=\mat{P}.
    \end{aligned}
    \end{equation*}

    \item Probabilistic robust matrix factorization (PRMF) \cite{wang2012probabilistic}:
    \begin{equation*}
    \begin{aligned}
        &\underset{\mat{U,V}}{\min}~ \|\mat{M-UV}^T\|_1+\frac{\lambda_u}{2\lambda}\|\mat{U}\|_2^2+\frac{\lambda_v}{2\lambda}\|\mat{V}\|_2^2,\\
        & u_{ij}|\lambda_u \sim \mathcal{N}(u_{ij}|~\mat{0},\lambda_u^{-1}),~v_{ij}|\lambda_u \sim \mathcal{N}(v_{ij}|~\mat{0},\lambda_v^{-1}).
    \end{aligned}
    \end{equation*}

\end{itemize}

The above four models are all based on low-rank assumptions to detect anomalies. The vanilla RPCA model applies matrix nuclear norm (convex surrogate) to approximate the matrix rank while RPCA-TV and PRCA-$\tau$ use matrix factorization (nonconvex surrogate) instead. Unlike RPCA, RPCA-TV and PRCA-$\tau$ consider the temporal correlation underlying the data. PRMF is also based on matrix factorization but from a probabilistic view, making the model have anomaly detection and completion ability. The Greece lowercase letter denotes the hyper-parameters in the models, which needs to be carefully tuned in the experiment.

\subsection{Synthetic Experiment}

\subsubsection{Synthetic data and measurement}
The anomaly detection performance is usually hard to measure as the ground truth of anomaly is unavailable in practice. In light of this, we first design synthetic periodic time-series data to compare the performance between the proposed method and the baseline models. The corrupted synthetic data $\mat{M}_\text{syn} \in \mathbb{R}^{N \times T}$ is given by
\begin{equation}
    \mat{M}_\text{syn} = \mat{L}_\text{syn} + \mat{S}_\text{syn} + \epsilon_{ij},
\end{equation}
where $\mat{L}_\text{syn}$ is a periodic low-rank matrix, $\mat{S}_\text{syn}$ is a sparse matrix, and the Gaussian noise $\epsilon_{ij}\sim \mathcal{N}(0,\sigma_\text{noise}^2), ~i \in [1,\cdots,N], ~j \in [1, \cdots, T]$ \cite{vaswani2018static}.

To obtain the periodic low-rank matrix, we generate two smaller matrices $\mat{U}_\text{syn} \in \mathbb{R}^{N\times R}$ and $\mat{V}_\text{syn} \in \mathbb{R}^{R\times T}$ as following:
\begin{equation}
\begin{aligned}
    &[\mat{U}_\text{syn}]_{i,r} \sim \mathcal{N}(0,\sigma_U^2),\\
    &[\mat{V}_\text{syn}]_{r,:} = \sin{\frac{\pi}{4}rt+\frac{\pi}{4}r},
\end{aligned}
\end{equation}
where $r \in [1,\dots,R]$ and $t \in [0.1,0.2,\dots, T/10]$. Therefore, the $\mat{L}_\text{syn}$ can be given by $\mat{L}_\text{syn}=\mat{U}_\text{syn}\mat{V}_\text{syn}$ with rank $R$.

To generate the sparse anomalies in the data, we first randomly sample the index set $\Lambda$ to indicate the location of anomaly and then we randomly sample the magnitude of anomalies:
\begin{equation}
    \mathcal{P}_\Lambda (\mat{S}_\text{syn})_{i,j} \sim \mathcal{N}(0,\sigma_S^2).
\end{equation}

 %


As $\mat{S}_\text{syn}$ is the ground-truth of anomalies, we use the mean absolute error (MAE) and root mean square error (RMSE) to assess the anomaly detection performance from full or partial observations:
\begin{equation}
\begin{aligned}
\label{eq:rmse}
    &\text{MAE} = \frac{1}{|\Omega|}\left\|\mathcal{P}_\Omega(\mat{S}_\text{syn}-\hat{\mat{S}}_\text{syn})\right\|_1,\\
    &\text{RMSE} = \sqrt{\frac{1}{|\Omega|}\|\mathcal{P}_\Omega(\mat{S}_\text{syn}-\hat{\mat{S}}_\text{syn})\|_F^2},
\end{aligned}
\end{equation}
where $\hat{\mat{S}}_\text{syn}$ is the estimated sparse matrix from the models and $|\Omega|$ is the number of the observations.

\subsubsection{Experiment setup}
\begin{figure}[htpb]
    \centering
    \includegraphics{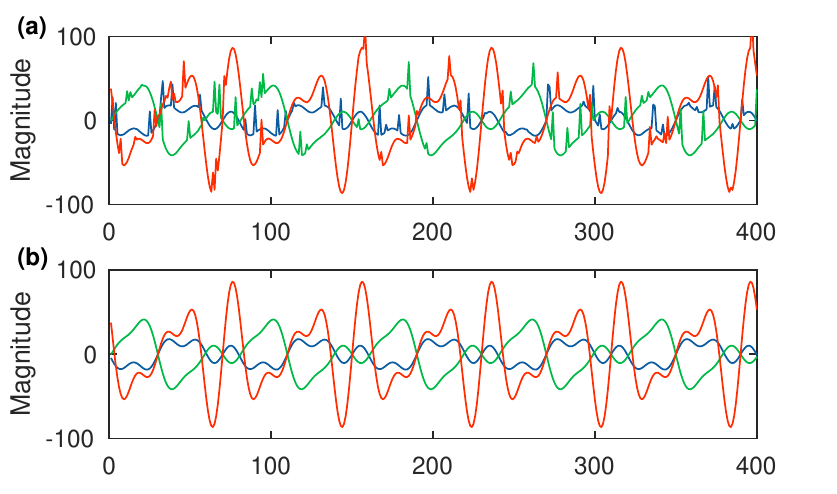}
    \caption{The generated multivariate time-series in synthetic data: (a) the first three corrupted time-series $[\mat{M}_\text{syn}]_{1:3,1:400}$; (b) the first three time-series in low-rank matrix $[\mat{L}_\text{syn}]_{1:3,1:400}$.}
    \label{fig:syn}
\end{figure}

In the experiment, we set $N=100$, $T=1200$, $R=4$, $\sigma_\text{noise}=0.1$,  $\sigma_{U}=20$, and $\sigma_S=40$ to generate the corrupted synthetic multivariate time-series. Figure \ref{fig:syn} shows the first three corrupted time-series for $t=1,\cdots,400$ and its corresponding low-rank time-series. There are 10\% entries as anomalies in $\mat{M}_\text{syn}$.

 Hyper-parameters are crucial to models and need to be carefully tuned. We set $\lambda= 0.05 ~(1/\sqrt{\max(N,T)})$ in RPCA \cite{candes2011robust}, $\lambda_1=10, \lambda_2=0,1$ and $\text{rank}=4$ in RPCA-TV and RPCA-$\tau$ model. Moreover, $\tau=80$ for RPCA-$\tau$ model as the periodic of the synthetic data is 80. In the PRMF model, $\lambda_u$ and $\lambda_v$ are both 5 and the $\text{rank}=4$. In the proposed model, we set $\lambda=0.002$ and the delay embedding length $\tau=80$. The $\beta$ and $\rho$ are 1.1 and $5\times 10^{-5}$ for RPCA-based models which apply the ADMM framework to solve the optimization problem. The same stopping criterion \textit{tol} $ = 1\times 10^{-5}$ is used for all the models.

\subsubsection{Results}

To quantify the anomaly detection performance of the models, the MAE and RMSE defined in \eqref{eq:rmse} when $\Omega = |NT|$ are given in Table \ref{Tab:syn}. Thanks to the Hankel structure, the proposed HT-RPCA exhibits the minimal MAE and RMSE in sparse matrix estimation. It demonstrates the proposed method has a better performance on anomaly detection task. We also found that the vanilla RPCA model shows a slightly better performance than RPCA-TV and RPCA-$\tau$ with Toeplitz temporal constraint. In other words, the temporal constraint might degrade the model performance if it cannot properly capture the temporal correlation underlying the data, such as the data includes small noise. Moreover, it has to be noted that we set the true rank ($R=4$) for RPCA-TV, RPCA-$\tau$, and RPMF model to obtain the results. However, the matrix rank is usually  prior knowledge in practice, which needs to be tuned according to the specific data.

\begin{table}[!h]
\centering
\begin{threeparttable}
\begin{tabular}{@{}llllll@{}}
\toprule
     & RPCA   & RPCA-TV & RPCA-$\tau$ & PRMF\tnote{1}   & HT-RPCA \\ \midrule
RMSE &  0.0963      & 0.0997         & 0.0999          & 0.1004 & \textbf{0.0772}    \\
MAE & 0.0710 & 0.0775  & 0.0776   & 0.0782  & \textbf{0.0490}          \\ \bottomrule
\end{tabular}
     \begin{tablenotes}
     \item[1] MAE and RMSE are the average value over 10 runs.
   \end{tablenotes}
\end{threeparttable}

\caption{Performance comparison (MAE/RMSE) between 4 baseline models and HT-RPCA for anomaly detection on synthetic data.}
\label{Tab:syn}
\end{table}

\begin{figure}[!htpb]
    \centering
    \includegraphics{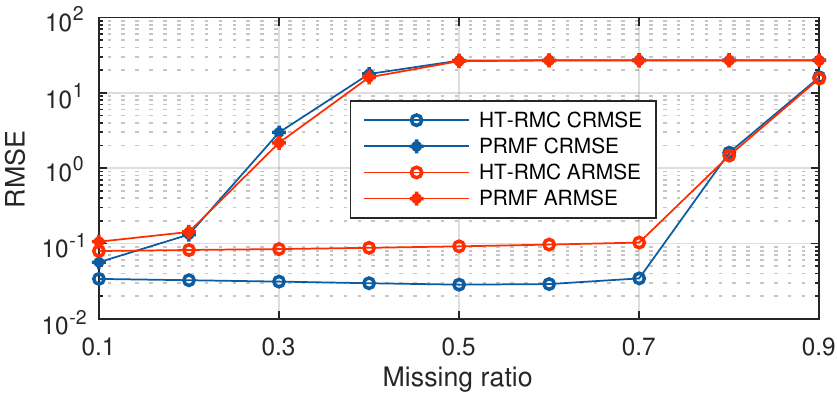}
    \caption{Performance comparison between PRMF and HT-RMC for anomaly detection and completion under different missing rate.}
    \label{fig:syn_miss}
\end{figure}

Furthermore, we compare the proposed HT-RMC model with the PRMF to measure the anomaly detection performance and the completion/imputation performance under different missing data scenarios. In the experiments, we randomly remove a certain amount of observations in $\mat{M}_\text{syn}$ to generate the index set $\Omega$ under different missing ratios. The results are shown in Figure \ref{fig:syn_miss}, in which CRMSE and ARMSE represent RMSE on completion task and anomaly detection task, respectively. As we can see, the proposed method exhibits superior performance compared with RPMF in both anomaly detection and completion tasks when the missing ratio is less than 0.8. However, when the data has a high missing ratio, e.g., 0.9, both methods fail to achieve the tasks as they dissatisfy the incoherence assumption, which prevents the low-rank matrix $\mat{L}$ to be sparse \cite{candes2011robust,zhang2019correction}.

\subsection{Real-world Traffic Experiment}

\subsubsection{Metro passenger flow and low-rank analysis}
The boarding passenger flow is collected from Guangzhou metro, registering 15-minute-level boarding flow for 159 stations in July 2017. The spatiotemporal data can be regarded as multivariate time series and organized as a matrix $\mat{M}\in \mathbb{R}^{159\times (72\times20)}$ (we have 72 windows per day from 6:00 a.m. - 12:00 a.m. and 20 days without considering weekends).

\begin{figure}[!htpb]
    \centering
    \includegraphics{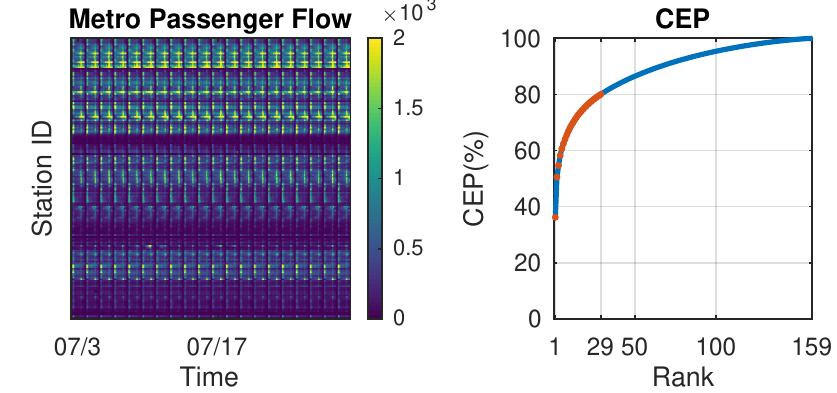}
    \caption{Left: The passenger boarding flow at each station. Right: The cumulative eigenvalue percentage of passenger boarding flow.}
    \label{fig:svd}
\end{figure}

The foundation of the proposed method and the baseline models is that the corrupted matrix $\mat{M}$ shows low-rank characteristics. Therefore, we use singular vector decomposition (SVD) to analyze the raw data at first. The raw passenger flow and its cumulative eigenvalue percentage (CEP) obtained from SVD are shown in Figure \ref{fig:svd}. We can see that the passenger flow exhibits clear periodic patterns, and the first 29 singular values can achieve 80\% CEP. In other words, the traffic data can be captured with the first few latent factors since most of the singular values are relatively small, which shows the low-rank property. The low-rank characteristic of the traffic data can also be extended to the Hankel tensor as the transformed Hankel tensor shows smooth manifolds in the low-rank space \cite{yokota2018missing}.

\subsubsection{Anomaly Detection}

Due to the highly heterogeneous passenger flow data in the spatial and temporal dimensions, it is challenging to set a global anomaly standard and measure the effect of anomalies in metro passenger flow \cite{wang2021diagnosing}. In addition, we do not have access to the ground-truth information of anomaly for this data set. Therefore, we only consider the increase/decrease in passenger flow except for the periodical fluctuation as anomalies in this work. We define that an anomaly occurs in station $n$ at time $t$ on day $k$ when
\begin{equation}
\label{eq:ano}
    M_{n,t}^k > {\bar{M}}_{n,t} + \xi\sigma_{n,t}, \quad \text{or} \quad {M}_{n,t}^k < {\bar{M}}_{n,t} -
    \xi\sigma_{n,t},
\end{equation}
where ${\bar{M}}_{n,t}=\sum_{k=1}^K M_{n,t}^k/K$ is the average passenger boarding flow in station $n$ at time $t$ during $K$ days, $\sigma_{n,t}$ is its standard derivation, and $\xi$ is a parameter to control the standard derivation.

\begin{figure*}[!t]
    \centering
    \includegraphics{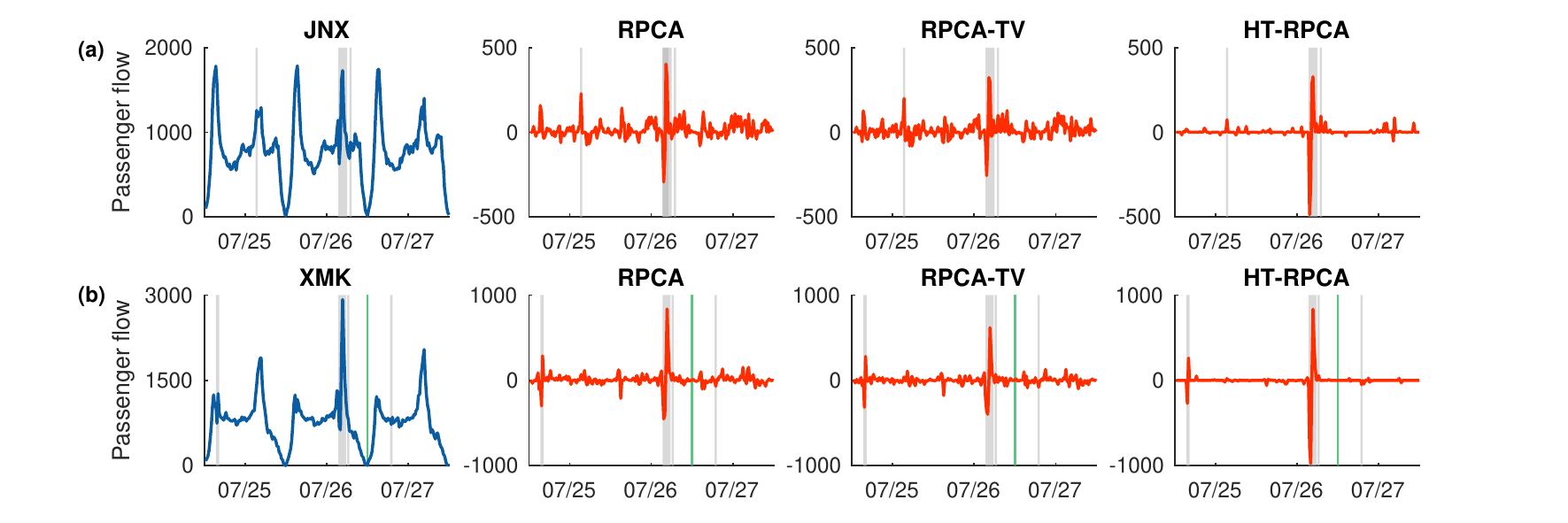}
    \caption{The anomaly detection results in metro passenger flow data. Blue line: the original passenger flow; red line: the anomaly obtained from the models. The gray patches represent the anomaly occurrence defined in \eqref{eq:ano} when $\xi=2$.}
    \label{fig:anomaly}
\end{figure*}

In the real-world traffic data experiment, we set $\gamma=0.0264$ in RPCA and $\tau = 72, \lambda_1=0.1$ and $\lambda_2=1\times 10^{-3}$ in RPCA-TV. In the HT-RPCA model, the hyperparameters $\tau$ and $\gamma$ are set as 72 and 0.0264, respectively. We show the original passenger flow (blue line) and anomaly passenger flow (red line) from the sparse matrix $\mat{S}$ of two selected stations (JNX and XMK) using three RPCA-based models in Figure~\ref{fig:anomaly}. The gray patches represent the anomaly occurrence defined in \eqref{eq:ano} when $\xi=2$.

In the original passenger data, the flows exhibit apparently recurrent daily patterns except for the gray patch areas, where the passenger flows change dramatically. We can get two conclusions from Figure~\ref{fig:anomaly}. (1) The three models can detect almost all anomalies except the one in green patch (23:45 July 26th in XMK station). Because when passenger flow is small, e.g., 4 passengers in this case, it is inaccurate to use statistical measurements, such as Eq.~\eqref{eq:ano}, to detect anomalies. (2) The anomaly passenger flow obtained from the RPCA and the RPCA-TV models shows more variations and fluctuations than the proposed method. Namely, the proposed method can accurately detect anomalies with fewer false alarms. It is the advantage of incorporating temporal Hankel delay embedding structure to capture more correlations from higher-dimension, such as the periodic patterns within the data.


\subsubsection{Anomaly propagation}
\begin{figure}[!t]
    \centering
    \includegraphics{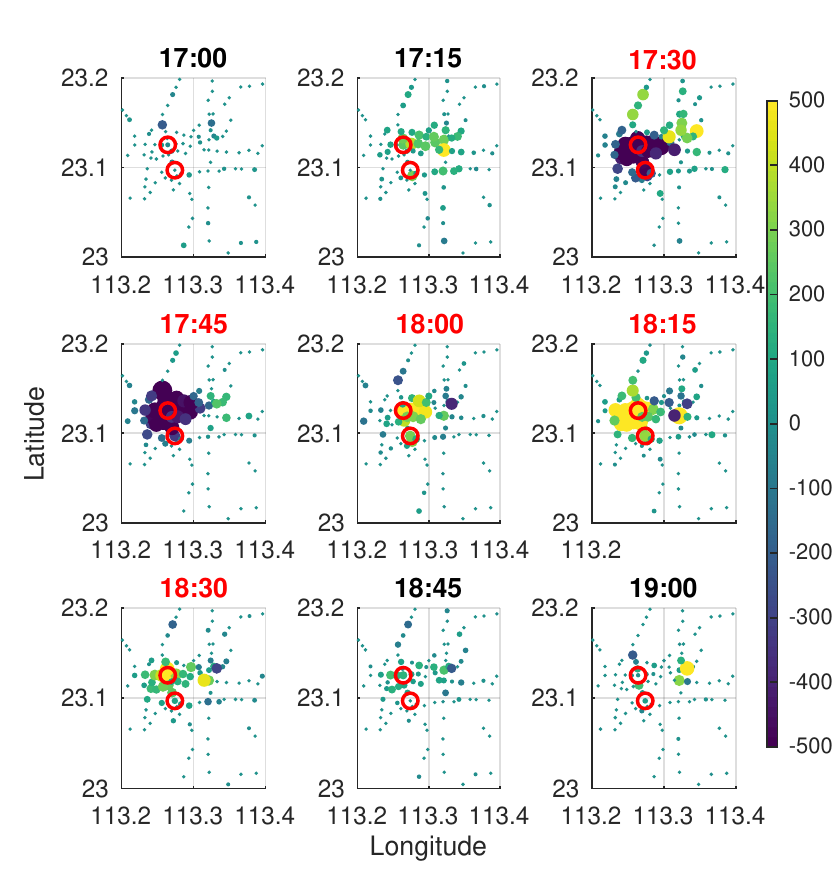}
    \caption{The anomaly distribution on July 26th. The red circles are station JNX and station XMK on the figure and the red title represents anomaly timestamp. The color of dots represents passenger flow.}
    \label{fig:ano_dist}
\end{figure}

We observe that station JNX and station XMK exhibit similar anomaly patterns, starting from 17:30 and ending at 18:30 on July 26th in Figure \ref{fig:anomaly}. In light of this, we further analyze the anomaly propagation through the metro network based on the abnormal passenger flow, i.e., $\mat{S}$. We show the anomalies at each station from 17:00 to 19:00 on July 26th in Figure \ref{fig:ano_dist}. The larger the dots are, the greater the absolute value of abnormal passenger flow is. We can see that before anomalies occurring in JNX and XMK, the boarding passenger flow of several stations near JNX and XMK slightly increased at 17:15. Then the boarding flow of some stations (purple dots), including JNX and XMK, suddenly decreases at 17:30. The phenomenon spreads to more stations along the same metro line in 15 minutes. After the short break, the boarding flow of affected stations starts to increase till 18:30. The anomaly has delayed occurrence between 18:00 and 18:15 at a few stations far from JNX and XMK. The entire anomaly duration lasts 1 hour, and it corresponds to the raw boarding flow of JNX and XMK in Figure \ref{fig:anomaly} (blue line).

\section{Conclusion and Discussion} \label{sec:conclusion}

In this study, we propose a new RPCA-based model for anomaly detection in spatiotemporal traffic data (e.g., traffic flow), which can be formulated as a multivariate time-series matrix. The basic principle of RPCA is to decompose a corrupted matrix into a low-rank matrix and a sparse matrix. Given that anomaly is rare and very different from regular values, the sparse matrix therefore can be used to characterize anomalies. Unlike previous studies based on the matrix-based RPCA, we propose an enhanced tensor version of RPCA by incorporating temporal Hankel delay embedding to augment the corrupted data and detect anomalies. Specifically, we apply the Hankelization on the temporal domain of the matrix to obtain an augmented third-order Hankel tensor. In doing so, the model can capture more dependency/correlation (e.g., periodic information) underlying the data and make the model more robust consequently. Then we propose an efficient algorithm named HT-RPCA to solve the tensor extension RPCA problem by minimizing the weighted sum of TNN of the Hankel tensor and $l_1$ norm of the matrix. We use the ADMM framework to solve the optimization problem. A modified model HT-RMC is also proposed to detect anomalies for data with missing values. We conduct two experiments on time series of synthetic data and metro passenger boarding flow collected from Guangzhou, China. The results verify the effectiveness and superiority of the proposed framework for anomaly detection.

There are some directions for future work. First, we can analyze and incorporate the causality of the anomaly to a forecasting model to make predictions more accurate. Second, computing SVD in updating the tensor nuclear norm at each iteration causes the computational cost, in particular for large-scale data; therefore, we can apply faster SVD strategies or nonconvex methods to approximate the rank to speed up the algorithm. Third, the proposed method only incorporates the additional temporal constraint by Hankelization process. We can also leverage the spatial constraint, e.g., topology information, into the model. Fourth, since the current model is based on historical data to detect anomalies, an online anomaly detection model is worth studying to achieve  early warning.


\ifCLASSOPTIONcaptionsoff
  \newpage
\fi




\bibliographystyle{IEEEtranN}
\bibliography{ref}

\begin{IEEEbiography}[{\includegraphics[width=1in,height=1.25in,clip,keepaspectratio]{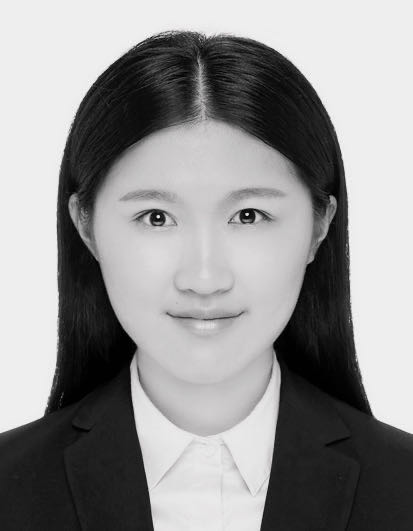}}]{Xudong Wang}
received the B.S. degree from Sichuan University, Sichuan, China, in 2014 and the M.S. degree from Beihang University, Beijing, China, in 2017. She is currently working toward the Ph.D. degree in the Department of Civil Engineering at McGill University, Montreal, QC, Canada. Her research interests include spatio-temporal traffic data mining and anomaly detection.
\end{IEEEbiography}

\begin{IEEEbiography}[{\includegraphics[width=1in,height=1.25in,clip,keepaspectratio]{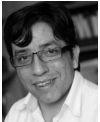}}]{Luis Miranda-Moreno} received his Ph.D. degree from University of Waterloo, Ontario, Canada. He is current an Associate Professor with the Department of Civil Engineering, McGill University. His research interests include the development of crash-risk analysis methods, the integration of emergency technologies for traffic monitoring, the impact of climate on transportation systems, the analysis of short and long-term changes in travel demand, the impact of transport on the environment, the evaluation of energy efficiency measures and non-motorized transportation.
\end{IEEEbiography}

\begin{IEEEbiography}[{\includegraphics[width=1in,height=1.25in,clip,keepaspectratio]{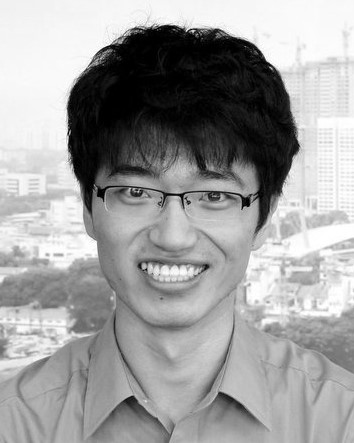}}]{Lijun Sun} (member, IEEE) received the B.S. degree in Civil Engineering from Tsinghua University, Beijing, China, in 2011, and Ph.D. degree in Civil Engineering (Transportation) from National University of Singapore in 2015. He is currently an Assistant
Professor with the Department of Civil Engineering at McGill University, Montreal, QC, Canada. His research centers on intelligent transportation systems, machine learning, spatiotemporal modeling, travel behavior, and agent-based simulation.
\end{IEEEbiography}

\end{document}